\documentclass{article}

\usepackage[utf8]{inputenc} 
\usepackage[T1]{fontenc}    
\usepackage{lmodern}
\usepackage{url}            
\usepackage{booktabs}       
\usepackage{amsfonts}       
\usepackage{nicefrac}       
\usepackage{microtype}      
\usepackage{setspace}
\usepackage{caption}
\usepackage{enumitem}
\usepackage{xfrac}
\usepackage{amsmath}
\usepackage[square,semicolon]{natbib}
\usepackage[backref=page]{hyperref}
\usepackage{xcolor}
\usepackage{rotating}
\usepackage[paper=8.5in:11in]{typearea}
\usepackage[margin=1.0in]{geometry}
\usepackage{amssymb}

\usepackage{multicol}
\usepackage{microtype}
\usepackage{graphicx}
\usepackage[procnames]{listings}
\definecolor{keywords}{RGB}{255,0,90}
\definecolor{comments}{RGB}{0,0,113}
\definecolor{red}{RGB}{160,0,0}
\definecolor{green}{RGB}{0,100,0}
 
\definecolor{darkblue}{rgb}{0, 0.2, 0.7}
\hypersetup{
    colorlinks = true,
    linkcolor = darkblue,
    anchorcolor = darkblue,
    citecolor = darkblue,
    filecolor = darkblue,
    urlcolor = darkblue
}

\title{\vspace{-2em}%
  \hrule height 4pt%
  \vskip 0.25in%
  \vskip -\parskip%
  \textbf{
  Fast Transformer Decoding: One Write-Head is All You Need
  }%
  \vskip 0.2in%
  \vskip -\parskip%
  \hrule height 1pt%
  \vskip 0.09in}

\author{Noam Shazeer \\ Google \\ noam@google.com}

\begin{document}

\maketitle

\begin{abstract}

Multi-head attention layers, as used in the Transformer neural sequence model, are a powerful alternative to RNNs for moving information across and between sequences.  While training these layers is generally fast and simple, due to parallelizability across the length of the sequence, incremental inference (where such paralleization is impossible) is often slow, due to the memory-bandwidth cost of repeatedly loading the large "keys" and "values" tensors. We propose a variant called multi-query attention, where the keys and values are shared across all of the different attention "heads", greatly reducing the size of these tensors and hence the memory bandwidth requirements of incremental decoding.  We verify experimentally that the resulting models can indeed be much faster to decode, and incur only minor quality degradation from the baseline.

\end{abstract}


\section{Introduction}
The Transformer neural sequence model \citep{Vas17} has emerged as a popular alternative to recurrent sequence models.  Transformer relies on attention layers to communicate information between and across sequences.  One major challenge with Transformer is the speed of incremental inference.  As we will discuss, the speed of incremental Transformer inference on modern computing hardware is limited by the memory bandwidth necessary to reload the large "keys" and "values" tensors which encode the state of the attention layers.  In the following sections, we will review the multi-head-attention layers used by Transformer, provide a performance analysis, and propose an architectural variation (multi-query attention) which greatly improves inference speed with only minor quality degradation.

\section{Background: Neural Attention}

Neural Attention, introduced by \citep{1409.0473}, is a powerful tool for manipulating variable-length representations.  A neural attention function takes a single query-vector $q$ and a set of $m$ different (key-vector, value-vector) pairs (represented by the matrices $K$ and $V$), and produces an output vector $y$.  The output $y$ is computed as a weighted sum of the different value vectors, where the weights are derived by comparing the query to the keys.

\subsection{Dot-Product Attention}

The following code describes a common formulation, where the weights are computed as the softmax of the dot-products of the query with the different keys.

\begin{minipage}{\linewidth}
\begin{lstlisting}[language=Python]
def DotProductAttention(q, K, V):
  """Dot-Product Attention on one query.
  Args:
    q: a vector with shape [k]
    K: a matrix with shape [m, k]
    V: a matrix with shape [m, v]
  Returns:
    y: a vector with shape [v]
  """
  logits = tf.einsum("k,mk->m", q, K)
  weights = tf.softmax(logits)
  return tf.einsum("m,mv->v", weights, V)
\end{lstlisting}
\end{minipage}

Our code samples use \textbf{einsum} notation, as defined in TensorFlow and numpy, for generalized contractions between tensors of arbitrary dimension.  In this notation, an equation names the dimensions of the input and output Tensors.  The computation is numerically equivalent to broadcasting each input to have the union of all dimensions, multiplying component-wise, and summing across all dimensions not in the desired output shape.

\subsection{Multi-head Attention}

The "Transformer" seuqence-to-sequence model \citep{Vas17} uses $h$ different attention layers (heads) in parallel, which the authors refer to as "Multi-head attention".  The query vectors for the $h$ different layers are derived from $h$ different learned linear projections $P_q$ of an input vector $x$.  Similarly, the keys and values are derived from $h$ different learned linear projections $P_k, P_v$ of a collection $M$ of $m$ different input vectors.  The outputs of the $h$ layers are themselves passed through different learned linear projections $P_o$, then summed.  For simplicity, we give the input and output vectors identical dimensionality $d$.  The The computation can be expressed as follows:

\begin{minipage}{\linewidth}
\begin{lstlisting}[language=Python]
def MultiheadAttention(
    x, M, P_q, P_k, P_v, P_o):
  """Multi-head Attention on one query.
  Args:
    x: a vector with shape [d]
    M: a matrix with shape [m, d]
    P_q: a tensor with shape [h, d, k]
    P_k: a tensor with shape [h, d, k]
    P_v: a tensor with shape [h, d, v]
    P_o: a tensor with shape [h, d, v]
  Returns:
    y: a vector with shape [d]
  """
  q = tf.einsum("d,hdk->hk", x, P_q)
  K = tf.einsum("md,hdk->hmk", M, P_k)
  V = tf.einsum("md,hdv->hmv", M, P_v)
  logits = tf.einsum("hk,hmk->hm", q, K)
  weights = tf.softmax(logits)
  o = tf.einsum("hm,hmv->hv", weights, V)
  y = tf.einsum("hv,hdv->d", o, P_o)
  return y
\end{lstlisting}
\end{minipage}

Note: \citep{Vas17} include a constant scaling factor on the logits.  We omit this in our code, as it can be folded into the linear projections $P_q$ or $P_k$.  

\subsection{Multi-head Attention (Batched)} \label{sec:mhab}
In practice, it is far more efficient to batch together multiple queries.  The code below adds two types of batching.  First, we generate queries from $n$ different positions in a sequence.  These queries all interact with the same keys and values.   In addition, we process a batch of $b$ different non-interacting sequences at once.  Following \citep{Vas17}, in an autoregressive model, we can prevent backward-information-flow by adding a "mask" to the logits containing the value $-\infty$ in the illegal positions.  

\begin{minipage}{\linewidth}
\begin{lstlisting}[language=Python]
def MultiheadAttentionBatched(
    X, M, mask, P_q, P_k, P_v, P_o):
  """Multi-head Attention.
  Args:
    X: a tensor with shape [b, n, d]
    M: a tensor with shape [b, m, d]
    mask: a tensor with shape [b, h, n, m]
    P_q: a tensor with shape [h, d, k]
    P_k: a tensor with shape [h, d, k]
    P_v: a tensor with shape [h, d, v]
    P_o: a tensor with shape [h, d, v]
  Returns:
    Y: a tensor with shape [b, n, d]
  """
  Q = tf.einsum("bnd,hdk->bhnk", X, P_q)
  K = tf.einsum("bmd,hdk->bhmk", M, P_k)
  V = tf.einsum("bmd,hdv->bhmv", M, P_v)
  logits = tf.einsum("bhnk,bhmk->bhnm", Q, K)
  weights = tf.softmax(logits + mask)
  O = tf.einsum("bhnm,bhmv->bhnv", weights, V)
  Y = tf.einsum("bhnv,hdv->bnd", O, P_o)
  return Y
\end{lstlisting}
\end{minipage}

\subsubsection{Performance Analysis of Batched Multi-head Attention} \label{sec:assu}
To simplify the performance analysis, we will make several simplifying assumptions:
\begin{itemize}
    \item $m = n$
    \item $k = v = \frac{d}{h}$, as suggested by \citep{Vas17}
    \item $n \leq d$
\end{itemize}

The total number of arithmetic operations is $\Theta(bnd^2)$.  (Since the complexity of each of the \texttt{tf.einsum} operations above is $O(bnd^2)$ given the simplifying assumptions.

The total size of memory to be accessed is equal to the sum of the sizes of all the tensors involved: $O(bnd + bhn^2 + d^2)$.  The first term is due to $X$, $M$, $Q$, $K$, $V$, $O$ and $Y$, the second term due to the logits and weights, and the third term due to the projection tensors $P_q$, $P_k$, $P_v$ and $P_o$.

Dividing the two, we find that the ratio of memory access to arithmetic operations is $O(\frac{1}{k} + \frac{1}{bn})$.  This low ratio is necessary for good performance on modern GPU/TPU hardware, where the computational capacity can be two orders of magnitude higher than the memory bandwidth.

\subsection{Multihead Attention (Incremental)}

In some settings, data dependencies make it is impossible to process queries from multiple positions in parallel.  An example is a self-attention layer in an autoregressive language model such as Transformer \citep{Vas17}.  The queries produced at each position attend to key-value pairs produced at all positions up to and including that position.  During training, the ground-truth target sequence is known, and we can use an efficient parallel implementation similar to that in section \ref{sec:mhab}.  However, when generating from the trained model, the output of the self-attention layer at a particular position affects the token that is generated at the next position, which in turn affects the input to that layer at the next position.   This prevents parallel computation.  Code for incrementally computing this self-attention layer is shown below.

\begin{minipage}{\linewidth}
\begin{lstlisting}[language=Python]
def MultiheadSelfAttentionIncremental(
    x, prev_K, prev_V, P_q, P_k, P_v, P_o):
  """Multi-head Self-Attention (one step).
  Args:
    x: a tensor with shape [b, d]
    prev_K: tensor with shape [b, h, m, k]
    prev_V: tensor with shape [b, h, m, v]
    P_q: a tensor with shape [h, d, k]
    P_k: a tensor with shape [h, d, k]
    P_v: a tensor with shape [h, d, v]
    P_o: a tensor with shape [h, d, v]
  Returns:
    y: a tensor with shape [b, d]
    new_K: tensor with shape [b, h, m+1, k]
    new_V: tensor with shape [b, h, m+1, v]
  """
  q = tf.einsum("bd,hdk->bhk", x, P_q)
  new_K = tf.concat(
    [prev_K, tf.expand_dims(tf.einsum("bd,hdk->bhk", M, P_k),axis=2)],
    axis=2)
  new_V = tf.concat(
    [prev_V, tf.expand_dims(tf.einsum("bd,hdv->bhv", M, P_v), axis=2)],
    axis=2)
  logits = tf.einsum("bhk,bhmk->bhm", q, new_K)
  weights = tf.softmax(logits)
  o = tf.einsum("bhm,bhmv->bhv", weights, new_V)
  y = tf.einsum("bhv,hdv->bd", O, P_o)
  return y, new_K, new_V
\end{lstlisting}
\end{minipage}

\subsubsection{Performance Analysis}

We make the same simplifying assumptions as in section \ref{sec:assu}.

Across $n$ calls, the total number of arithmetic operations is again $\Theta(bnd^2)$.

Across $n$ calls, the total amount of memory access is $\Theta(bn^2d + nd^2)$, the first term due to $K$ and $V$ and the second term due to  $P_q$, $P_k$, $P_v$ and $P_o$.

Dividing the memory by the computations, we find that the ratio of memory access to arithmetic operations is $\Theta(\frac{n}{d} + \frac{1}{b})$.  When $n \approx d$ or $b \approx 1$, the ratio is close to 1, causing memory bandwidth to be a major performance bottleneck on modern computing hardware.  In order to make incremental generation efficient, we must reduce both of these terms to be $\ll1$.  The $\frac{1}{b}$ term is the easier one - we can just use a larger batch size, memory size permitting.

Reducing the $\frac{n}{d}$ term is harder.  This term is related to the expense of reloading at each step the $K$ and $V$ tensors representing the memory which have size $bhmk=bn^2$. One solution is to limit the sequence length $n$.  Another is to reduce the number of positions being attended-to, either by attending to a local neighborhood, or by otherwise compressing the number of memory positions,  as in \citep{liu2018generatin}, \citep{1805.00631}, \citep{povey2018time}.  In this paper we present an orthogonal approach to reducing the size of the $K$ and $V$ tensors - namely removing their "heads" dimension, while maintaining the "heads" dimension in the queries.

\section{Multi-Query Attention}

We introduce \textbf{multi-query Attention} as a variation of multi-head attention as described in \citep{Vas17}.  Multi-head attention consists of multiple attention layers (heads) in parallel with different linear transformations on the queries, keys, values and outputs.  Multi-query attention is identical except that the different heads share a single set of keys and values.  The code for (incremental) multi-query (self) attention is identical to the code listed above for multi-head attention, except that we remove the letter "h" from the \texttt{tf.einsum} equations where it represents the "heads" dimension of $K$, $V$, $P_k$, or $P_v$.

\begin{minipage}{\linewidth}
\begin{lstlisting}[language=Python]
def MultiqueryAttentionBatched(
    X, M, mask, P_q, P_k, P_v, P_o):
  """Multi-Query Attention.
  Args:
    X: a tensor with shape [b, n, d]
    M: a tensor with shape [b, m, d]
    mask: a tensor with shape [b, h, n, m]
    P_q: a tensor with shape [h, d, k]
    P_k: a tensor with shape [d, k]
    P_v: a tensor with shape d, v]
    P_o: a tensor with shape [h, d, v]
  Returns:
    Y: a tensor with shape [b, n, d]
  """
  Q = tf.einsum("bnd,hdk->bhnk", X, P_q)
  K = tf.einsum("bmd,dk->bmk", M, P_k)
  V = tf.einsum("bmd,dv->bmv", M, P_v)
  logits = tf.einsum("bhnk,bmk->bhnm", Q, K)
  weights = tf.softmax(logits + mask)
  O = tf.einsum("bhnm,bmv->bhnv", weights, V)
  Y = tf.einsum("bhnv,hdv->bnd", O, P_o)
  return Y
\end{lstlisting}
\end{minipage}

\begin{minipage}{\linewidth}
\begin{lstlisting}[language=Python]
def MultiquerySelfAttentionIncremental(
    x, prev_K, prev_V, P_q, P_k, P_v, P_o):
  """Multi-query Self-Attention (one step).
  Args:
    x: a tensor with shape [b, d]
    prev_K: tensor with shape [b, m, k]
    prev_V: tensor with shape [b, m, v]
    P_q: a tensor with shape [h, d, k]
    P_k: a tensor with shape [d, k]
    P_v: a tensor with shape [d, v]
    P_o: a tensor with shape [h, d, v]
  Returns:
    y: a tensor with shape [b, d]
    new_K: tensor with shape [b, m+1, k]
    new_V: tensor with shape [b, m+1, v]
  """
   q = tf.einsum("bd,hdk->bhk", x, P_q)
  K = tf.concat(
    [prev_K, tf.expand_dims(tf.einsum("bd,dk->bk", M, P_k), axis=2)],
    axis=2)
  V = tf.concat(
    [prev_V, tf.expand_dims(tf.einsum("bd,dv->bv", M, P_v), axis=2)],
    axis=2)
  logits = tf.einsum("bhk,bmk->bhm", q, K)
  weights = tf.softmax(logits)
  o = tf.einsum("bhm,bmv->bhv", weights, V)
  y = tf.einsum("bhv,hdv->bd", O, P_o)
  return y, K, V
\end{lstlisting}
\end{minipage}

\subsection{Performance Analysis for Incremental Multi-Query Attention}

We make the same simplifying assumptions as in section \ref{sec:assu}.

Across $n$ calls, the total number of arithmetic operations is again $\Theta(bnd^2)$.

Across $n$ calls, the total amount of memory access is $\Theta(bnd + bn^2k + nd^2)$, the first term due to $x$, $q$, $o$ and $y$, the second term due to $K$ and $V$ and the third term due to $P_q$, $P_k$, $P_v$, $P_o$.

Dividing the memory by the computations, we find that the ratio of memory access to arithmetic operations is $\Theta(\frac{1}{d} + \frac{n}{dh} + \frac{1}{b})$.  We have reduced the offensive $\frac{n}{d}$ by a factor of $h$.  Theoretically, given large batch size $b$, this should dramatically improve performance of incremental generation.  In our experimental section, we will show that the performance gains are real and that model quality remains high.

\section{Experiments and Results}

\subsection{Experimental Setup}

Following \citep{Vas17}, we evaluate on the WMT 2014 English-German translation task.  As a baseline, we use an encoder-decoder Transformer model with 6 layers, using $d_{model}=1024$ $d_{ff}=4096$, $h=8$, $d_k=d_v=128$, learned positional embeddings, and weight-sharing between the token-embedding and output layers.  The baseline model and all variations have 211 million parameters.  All models were trained for 100,000 steps (~20 epochs).  Each training batch consisted of 128 examples, each of which consisted of a 256-token input sequence and a 256-token target sequence (multiple training sentences were concatenated together to reach this length).   Models were trained on a 32-core TPUv3 cluster, with each model taking about 2 hours to train.  We used an implementation from the tensor2tensor and mesh-tensorflow libraries.  The configurations used can be found at [to be added before publication]  , including details about learning rates, dropout, label smoothing, etc.

In our "multi-query" model, we replace all of the attention layers in the model to multi-query attention.  This includes the encoder-self-attention, decoder-self-attention and encoder-decoder-attention layers.  We widen the feed-forward hidden layers from 4096 to 5440 to make the total parameter-count equal to that of the baseline.

To demonstrate that local-attention and multi-query attention are orthogonal, we also trained "local" versions of the baseline and multi-query models, where the decoder-self-attention layers (but not the other attention layers) restrict attention to the current position and the previous 31 positions.  

A simpler alternative way to reduce the sizes of $K$ and $V$ is to reduce the number of heads $h$ and/or to reduce the dimensionalities $k$ and $v$ of the keys and values.  We trained several such models for comparison, again widening the feed-forward hidden layers to make the total parameter-count equal to that of the baseline.

We preformed a similar set of experiments using "transformer-decoder" language models on the Billion-Word Language Modeling Benchmark \citep{LM1B}.    For the baseline, we use a model with 6 layers, $d_{model}=1024$ $d_{ff}=8192$, $h=8$, $d_k=d_v=128$.  The total parameter count is 192 million for the baseline and for all variations.  We trained for 136K steps (10 epochs) at a batch size of 64K tokens.  Again, we used a 32-core TPUv3 cluster for approximately 3 hours to train each model.

\subsection{Model Quality}

Table \ref{tab:translation} shows results for the machine-translation experiments.  We decoded the dev set using greedy maximum-likelihood decoding and computed BLEU score with sacrebleu \texttt{"sacrebleu -t wmt13 -l en-de -tok intl"}.  We also list per-subword-token perplexity on the dev set.  According to both of these metrics, the multi-query attention model seems to be slightly worse than the baseline, but much closer than any of the alternatives involving decreasing $h$, $d_k$ and $d_v$. 

We validated the results by decoding the test set using both greedy decoding and beam search (beam 4, $\alpha=0.6$), and evaluated with sacrebleu \texttt{"sacrebleu -t wmt14 -l en-de -tok intl"}.  Again, the multi-query model performed similarly to the baseline, and actually had the highest BLEU score (28.5) with beam-4 decoding.

Table \ref{tab:lm1b} shows results for the billion-word language modeling benchmark.  Models were evaluated by per-word (not per-subword-token) perplexity on the dev set.  The results paint a similar picture to the translation results.  The multi-query attention model was slightly worse than the baseline, but significantly better than any of the alternatives involving decreasing $h$, $d_k$ and $d_v$.

\subsection{Speed}

Table \ref{tab:translation2} shows training and inference times for the various models.  Both training and inference speeds were evaluated on one TPUv2 (8 cores).  A training step (consisting of 32,768 input tokens and 32,768 target tokens, as described above) took 433ms for the base model and 425ms for the multi-query model.  Dividing by 32,768, we find that the training time is 13.2$\mu s$ per (input-token + target-token), as listed in Table \ref{tab:translation2}.

We ran incremental greedy inference on a batch of 1024 sequences (128 per core) using a source-sequence length of 128 tokens and a target sequence length of 128.   \footnote{Due to system limitations requiring fixed shapes, we used padding and masking in our decoder-self-attention implementation.  The memory tensors were thus padded to the maximum length (128), or to the window-size (32) in the case of local attention.  Each decoding step thus took the same amount of time. An alternative implementation of incrementally growing the tensors could save time near the beginning of the sequence.}   For the baseline model, the encoder part of the model took 222ms and each incremental step of the decoder took 47ms.  Dividing by the respective numbers of tokens, we find that the amortized inference time is $1.7\mu s$ per token for the encoder and a much larger $46\mu s$ per token for the decoder, as listed in Table \ref{tab:translation2}.   For the multi-query model, the encoder took 195ms and the decoder took 3.9ms per step, for amortized per-token costs of $1.5\mu s$ and $3.8\mu s$ respectively.  Table \ref{tab:translation2} shows these values as well as similar results for beam-search.

\begin{table}[h]
\caption{WMT14 EN-DE Results.}
\label{tab:translation}
\begin{center}
\begin{tabular}{cccc|cc|c}
\hline\rule{0pt}{2.0ex}
Attention & $h$ & $d_k, d_v$ & $d_{ff}$ &  ln(PPL) & BLEU & BLEU (test)\\
Type & & & & (dev) & (dev) & beam 1 / 4\\

\hline
multi-head & 8 & 128 & 4096 &  \textbf{1.424} & \textbf{26.7} & 27.7 / 28.4\\
multi-query & 8 & 128 & 5440 &  1.439 & 26.5 & 27.5 / \textbf{28.5}\\
multi-head local & 8 & 128 & 4096  & 1.427 & 26.6 & 27.5 / 28.3 \\
multi-query local & 8 & 128 & 5440 & 1.437 & 26.5 & 27.6 / 28.2 \\
\hline
multi-head & 1 & 128 & 6784 & 1.518 & 25.8 \\
multi-head & 2 & 64 & 6784 & 1.480 & 26.2 & 26.8 / 27.9\\
multi-head & 4 & 32 & 6784 & 1.488 & 26.1 \\
multi-head & 8 & 16 & 6784 & 1.513 &  25.8\\
\hline
\end{tabular}
\end{center}
\end{table}

\begin{table}[h]
\caption{Amortized training and inference costs for WMT14 EN-DE Translation Task with sequence length 128.  Values listed are in TPUv2-microseconds per output token.}
\label{tab:translation2}
\begin{center}
\begin{tabular}{r|c|c|c}
\hline\rule{0pt}{2.0ex}
Attention & Training & Inference  & Beam-4 Search \\
Type & & enc. + dec. & enc. + dec. \\

\hline
multi-head  & 13.2 & 1.7 + 46 & 2.0 + 203 \\
multi-query & \textbf{13.0} & 1.5 + 3.8 & 1.6 + 32 \\

multi-head local  & 13.2 & 1.7 + 23 & 1.9 + 47 \\
multi-query local & \textbf{13.0} & \textbf{1.5 + 3.3} & \textbf{1.6 + 16} \\
\hline
\end{tabular}
\end{center}
\end{table}

\begin{table}[h!]
\caption{Billion-Word LM Benchmark Results.}
\label{tab:lm1b}
\begin{center}
\scalebox{1.0}{
\begin{tabular}{cccc|c}
\hline\rule{0pt}{2.0ex}
Attention & $h$ & $d_k, d_v$ & $d_{ff}$  & dev-PPL \\
\hline
multi-head & 8 & 128 & 8192 &  \textbf{29.9} \\
multi-query & 8 & 128 & 9088 &  30.2 \\
\hline
multi-head & 1 & 128 & 9984 & 31.2 \\
multi-head & 2 & 64 & 9984 & 31.1 \\
multi-head & 4 & 32 & 9984 & 31.0 \\
multi-head & 8 & 16 & 9984 & 30.9 \\
\hline
\end{tabular}
}
\end{center}
\end{table}

\section{Conclusion}
We have proposed multi-query attention - an alternative to multi-head attention with much lower memory-bandwidth requirements in the incremental setting.  We believe that this enables wider adoption of attention-based sequence models in inference-performance-critical applications.

\bibliography{main}

\begin{thebibliography}{6}
\providecommand{\natexlab}[1]{#1}
\providecommand{\url}[1]{\texttt{#1}}
\expandafter\ifx\csname urlstyle\endcsname\relax
  \providecommand{\doi}[1]{doi: #1}\else
  \providecommand{\doi}{doi: \begingroup \urlstyle{rm}\Url}\fi

\bibitem[Bahdanau et~al.(2014)Bahdanau, Cho, and Bengio]{1409.0473}
Dzmitry Bahdanau, Kyunghyun Cho, and Yoshua Bengio.
\newblock Neural machine translation by jointly learning to align and
  translate, 2014.

\bibitem[Chelba et~al.(2013)Chelba, Mikolov, Schuster, Ge, Brants, and
  Koehn]{LM1B}
Ciprian Chelba, Tomas Mikolov, Mike Schuster, Qi~Ge, Thorsten Brants, and
  Phillipp Koehn.
\newblock One billion word benchmark for measuring progress in statistical
  language modeling.
\newblock \emph{CoRR}, abs/1312.3005, 2013.
\newblock URL \url{http://arxiv.org/abs/1312.3005}.

\bibitem[Liu et~al.(2018)Liu, Saleh, Pot, Goodrich, Sepassi, Kaiser, and
  Shazeer]{liu2018generatin}
Peter~J Liu, Mohammad Saleh, Etienne Pot, Ben Goodrich, Ryan Sepassi, Lukasz
  Kaiser, and Noam Shazeer.
\newblock Generating wikipedia by summarizing long sequences.
\newblock In \emph{Proceedings of the International Conference on Learning
  Representations}, 2018.

\bibitem[Povey et~al.(2018)Povey, Hadian, Ghahremani, Li, and
  Khudanpur]{povey2018time}
Daniel Povey, Hossein Hadian, Pegah Ghahremani, Ke~Li, and Sanjeev Khudanpur.
\newblock A time-restricted self-attention layer for {ASR}.
\newblock In \emph{Proceddings of the IEEE International Conference on
  Acoustics, Speech and Signal Processing (ICASSP)}. IEEE, 2018.

\bibitem[Vaswani et~al.(2017)Vaswani, Shazeer, Parmar, Uszkoreit, Jones, Gomez,
  Kaiser, and Polosukhin]{Vas17}
Ashish Vaswani, Noam Shazeer, Niki Parmar, Jakob Uszkoreit, Llion Jones,
  Aidan~N. Gomez, Lukasz Kaiser, and Illia Polosukhin.
\newblock Attention is all you need.
\newblock In \emph{NIPS}, 2017.

\bibitem[Zhang et~al.(2018)Zhang, Xiong, and Su]{1805.00631}
Biao Zhang, Deyi Xiong, and Jinsong Su.
\newblock Accelerating neural transformer via an average attention network,
  2018.

\end{thebibliography}
\bibliographystyle{plainnat}


\end{document}